\documentclass[10pt,twocolumn,letterpaper]{article}

\usepackage{iccv}              

\usepackage{algorithm}
\usepackage{algpseudocode}
\usepackage{booktabs}
\usepackage{threeparttable}
\usepackage[T1]{fontenc}
\usepackage{color}
\usepackage{xcolor}
\usepackage{multirow}
\usepackage{overpic}

\definecolor{iccvblue}{rgb}{0.21,0.49,0.74}
\usepackage[pagebackref,breaklinks,colorlinks,allcolors=iccvblue]{hyperref}

\def\paperID{10033} 
\def\confName{ICCV}
\def\confYear{2025}

\title{Agentic Keyframe Search for Video Question Answering}

\author{Sunqi Fan \quad Meng-Hao Guo \quad Shuojin Yang\\
Tsinghua University\\
{\tt\small \{fansq20, gmh20\}@mails.tsinghua.edu.cn,  yangshuojin@mail.tsinghua.edu.cn}
}

\begin{document}
\maketitle
\begin{abstract}

Video question answering (VideoQA) enables machines to extract and comprehend key information from 
videos through natural language interaction,
which is a critical step towards achieving intelligence.
However, the demand for a thorough understanding of videos and high computational costs still limit the widespread applications of VideoQA.
To address it,
we propose Agentic Keyframe Search (\textsc{AKeyS}),
a simple yet powerful algorithm 
for 
identifying keyframes in the VideoQA task. 
It can effectively distinguish key information from redundant, irrelevant content  
by leveraging modern language agents to direct classical search algorithms.
Specifically, we first segment the video and organize it as a tree structure.
Then, \textsc{AKeyS} uses a language agent to estimate heuristics and movement costs while dynamically expanding nodes. 
Finally, the agent determines if sufficient keyframes have been collected based on termination conditions and provides answers.
Extensive experiments on the EgoSchema and NExT-QA datasets show that \textsc{AKeyS} outperforms all previous methods with the highest keyframe searching efficiency, which means it can accurately identify key information and conduct effective visual reasoning with minimal computational overhead.
For example, on the EgoSchema subset, it achieves 1.8\% higher accuracy while processing only 43.5\% of the frames compared to VideoTree.
We believe that \textsc{AKeyS} represents a significant step towards building intelligent agents for video understanding.
The code is publicly available at \url{https://github.com/fansunqi/AKeyS}.

\end{abstract}    
\section{Introduction}
\label{sec:intro}

The rapid advancement of image-based Multimodal Large Language Models (MLLMs)~\cite{openai2024gpt4technicalreport, geminiteam2024geminifamilyhighlycapable} has significantly simplified image understanding tasks in daily life. 
Users can easily upload images to OpenAI’s GPT-4V or Google Gemini, ask questions about them, and 
receive responses via natural language interaction.
However, video understanding presents greater challenges, and the development of Video Large Language Models (Video-LLMs) has notably lagged behind image-based MLLMs.
Existing Video-LLMs often struggle to capture details in videos and lack a holistic understanding of video content~\cite{liu2024tempcompassvideollmsreally}. 
Moreover, the computational overhead of Video-LLMs is substantially higher than that of Large Language Models (LLMs) and image-based MLLMs, hindering their commercial deployment.
To address video understanding tasks in daily life more effectively, 
This paper focuses on the efficient extraction of keyframes, and analyzing them using image-based MLLMs for video understanding.


\begin{figure}[t]
  \centering 
  \footnotesize
  \begin{overpic}[width=\linewidth]{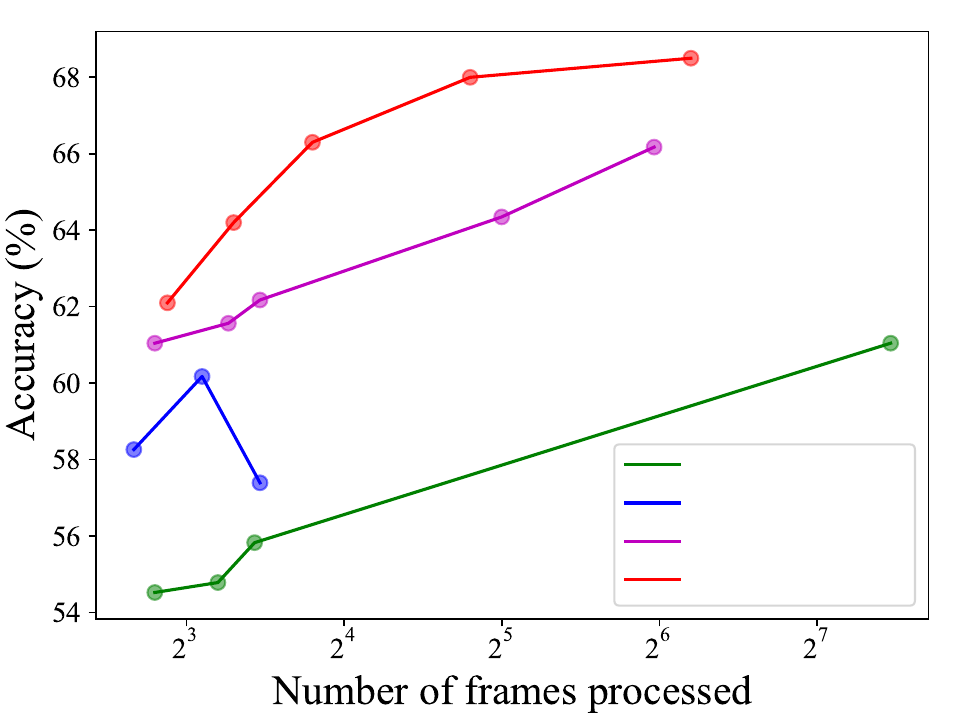}
    \put(72, 25.7){LLoVi~\cite{zhang2024simplellmframeworklongrange}}
    \put(72, 21.6){VideoAgent~\cite{wang2024videoagentlongformvideounderstanding}}
    \put(72, 17.5){VideoTree~\cite{wang2024videotreeadaptivetreebasedvideo}}
    \put(72, 13.7){AKeyS (Ours)}
  \end{overpic}
\caption{Demonstration of \textsc{AKeyS}'s high frame efficiency. When processing the same number of video frames with the same (M)LLM, \textsc{AKeyS} achieves higher QA accuracy. \textbf{At the same accuracy level (66\%), \textsc{AKeyS} uses only about 1/4 of the frames required by VideoTree.} Moreover, VideoTree clusters features of all frames during preprocessing, whereas \textsc{AKeyS} only has access to visible frames and does not utilize information from the rest. This experiment is conducted on EgoSchema~\cite{mangalam2023egoschemadiagnosticbenchmarklongform} subset.
}\label{fig:frame_efficiency}
\end{figure}

\begin{figure}[t]
  \centering
   \includegraphics[width=\linewidth]{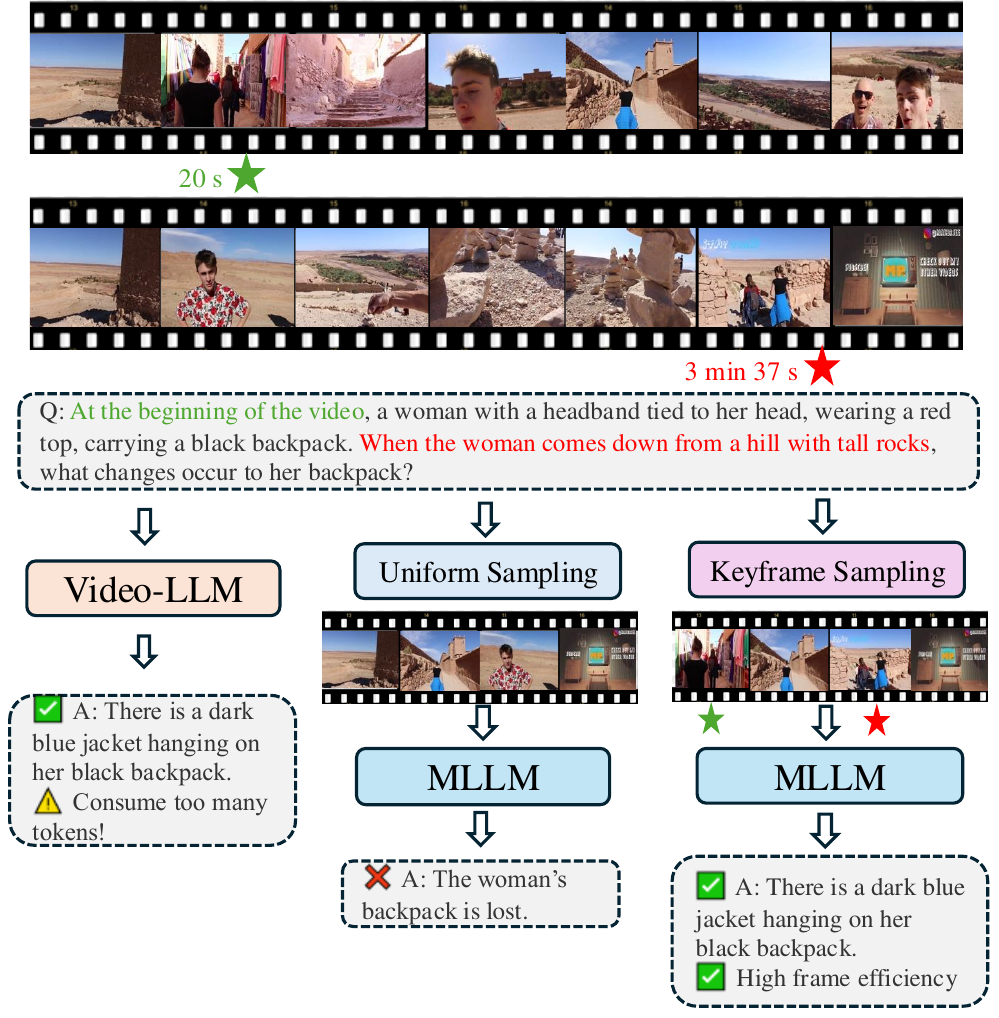}
   \caption{Comparison of three methods for analyzing a travel vlog: (1) Video-LLM can generate correct answers but is highly token-intensive; (2) The method of uniform frame sampling may introduce irrelevant content, leading MLLM to incorrect predictions; (3) The method of keyframe sampling for MLLM achieves both accuracy and efficiency. The keyframes relevant to the given question are highlighted in the figure.}
   \label{fig:keyframe}
\end{figure}

However, one key advantage of keyframe extraction is the ability to significantly reduce computational overhead while preserving essential information.
Figure \ref{fig:keyframe} presents three approaches to solving the VideoQA task. Among them, only the keyframe sampling based method achieves both accuracy and efficiency, highlighting the importance of keyframes in VideoQA task.
However, a major challenge is how to effectively identify keyframes that contain the essential information needed to answer specific questions.
This challenge becomes more pronounced in the context of long-form video understanding~\cite{mangalam2023egoschemadiagnosticbenchmarklongform, wu2021longformvideounderstanding}, where the abundance of irrelevant information necessitates precise temporal localization of key content based on the question at hand. 
Addressing efficiency and accuracy keyframe location challenge is crucial 
in video understanding tasks.

In this paper, we propose an efficient algorithm named \textsc{AKeyS} to tackle the video understanding and analysis problems, exemplified by VideoQA tasks.
Drawing inspiration from both traditional search algorithms and modern language agents, our approach harnesses the cognitive capabilities of language agents, such as reasoning, planning, summarization and reflection, to guide and provide feedback to traditional search algorithms.
This methodology effectively extracts key content from redundant information, similar to sifting wheat from chaff.

Specifically, given an input video, \textsc{AKeyS} divides it into segments and extracts textual information from the representative frame of each segment using a Vision-Language Model (VLM), such as an image captioner. 
Then, it employs language agents to perform temporal comparisons and identify key content in an iterative, deepening process until reaching the termination condition.
This leads to a tree-like search through the video until sufficient key information is found to answer the question.



In our experiments, \textsc{AKeyS} achieves 63.1\% accuracy on EgoSchema fullset~\cite{mangalam2023egoschemadiagnosticbenchmarklongform} (surpassing the best baseline by 2.0\%) and 77.4\% average accuracy on NExT-QA~\cite{xiao2021nextqanextphasequestionansweringexplaining} (surpassing the best baseline by 1.8\%).
In Figure \ref{fig:frame_efficiency}, we compare the frame efficiency of \textsc{AKeyS} with two baselines, highlighting its ability to effectively identify key information. In summary, \textsc{AKeyS} not only achieves state-of-the-art accuracy, but also exhibits the highest frame searching efficiency, making it a highly promising approach for real-world video analysis tasks.


\section{Related Work}
\label{sec:related work}

\textbf{Video Question Answering}~~VideoQA is a typical subtask of video understanding, involving the comprehension, analysis, and responding to questions about video content. 
It comprehensively tests various capabilities of multimodal QA systems~\cite{xiao2021nextqanextphasequestionansweringexplaining,zhong2022videoquestionansweringdatasets,nguyen2024videolanguageunderstandingsurveymodel}, while benchmarks and datasets for VideoQA have been progressively focused on longer videos and more complex reasoning scenarios~\cite{mangalam2023egoschemadiagnosticbenchmarklongform,fu2024videommefirstevercomprehensiveevaluation,wu2024longvideobenchbenchmarklongcontextinterleaved,zhou2025mlvubenchmarkingmultitasklong}. 

Early approaches for VideoQA typically employ neural networks (e.g., ResNet~\cite{he2015deepresiduallearningimage}, 3D convolutional neural networks~\cite{tran2015learningspatiotemporalfeatures3d,carreira2018quovadisactionrecognition,hara2018spatiotemporal3dcnnsretrace}) to extract visual features, while language models were used to process the questions. These two components are then aggregated to produce answers. With the advent of LLMs, the common practice is to use a pre-trained visual encoder to extract visual features, a projection layer to map visual representations into the text latent space of LLMs, and a pretrained LLM for response generation~\cite{zhang2023videollamainstructiontunedaudiovisuallanguage, wang2022internvideogeneralvideofoundation, lin2024videollavalearningunitedvisual,li2025temporalpreferenceoptimizationlongform}. These Video-LLMs, with their extensive parameters, can model long contexts~\cite{shen2024longvuspatiotemporaladaptivecompression, weng2024longvlmefficientlongvideo,song2024moviechatdensetokensparse} and, through instruction tuning and alignment, better address VideoQA tasks, serving as foundational models in the video domain. 
Another popular method is based on (M)LLMs or agents~\cite{xiao2024videoqaerallmsempirical,tang2024videounderstandinglargelanguage}.
Many works have achieved significant success by leveraging a range of agentic techniques, including prompting~\cite{zhang2024simplellmframeworklongrange, zhang2024hcqaego4degoschema}, memory~\cite{fan2024videoagentmemoryaugmentedmultimodalagent,kahatapitiya2024languagerepositorylongvideo, wang2024lifelongmemoryleveragingllmsanswering}, 
tools~\cite{choudhury2023zeroshotvideoquestionanswering, gupta2022visualprogrammingcompositionalvisual, yang2024doraemongptunderstandingdynamicscenes}, 
and planning~\cite{jeoung2024adaptivevideounderstandingagent, mahmood2024vurfgeneralpurposereasoningselfrefinement, min2024morevqaexploringmodularreasoning, ranasinghe2025understandinglongvideosmultimodal}. These advancements have progressively contributed to developing an intelligent and powerful video agent. A detailed discussion of these works can be found in the Appendix Section A.

\smallskip
\noindent
\textbf{Keyframe Extraction}~~Another line of research focuses on extracting keyframes from videos, which is also highly relevant to our work.
LVNet~\cite{park2024framesusefulefficientstrategies} selects keyframes using a small network that is specifically trained for this task. 
VCA~\cite{yang2024doraemongptunderstandingdynamicscenes} extracts key segments through selective attention. 
IG-VLM~\cite{kim2024imagegridworthvideo} performs uniform sampling across all frames, converting them into an image grid, which is then directly input into an off-the-shelf vision-language model. 
VideoAgent~\cite{wang2024videoagentlongformvideounderstanding} mimics the human brain's process by recursively selecting key video segments. After each selection, it uses LLM to evaluate whether there is enough confidence to answer the question based on the selected segments, terminating the recursion if sufficient confidence is reached.

VideoTree~\cite{wang2024videotreeadaptivetreebasedvideo} is the most closely related work to ours.  
VideoTree computes image features and performs k-means clustering on video segments, constructing a static video tree.
The LLM then searches along the tree until the key information is found and the question is answered. The key differences between our work and VideoTree are as follows:
(1) VideoTree computes image features for all video frames using CLIP~\cite{radford2021learningtransferablevisualmodels}, whereas \textsc{AKeyS} only extracts information from the visible frames. 
Therefore, \textsc{AKeyS} has lower computational overhead.
(2) VideoTree constructs a static video tree in advance, which remains unchanged regardless of the input question. In contrast, the video tree in \textsc{AKeyS} is dynamic and adaptive to the specific question, allowing us to identify the most suitable tree structure and optimal search path for each question.

\section{Method}
\label{sec:method}
\subsection{Background: Basic Searching Algorithms}

Our \textsc{AKeyS} algorithm is built on basic search algorithm in Algorithm \ref{algorithm_0}. Based on this fundamental process, the following search algorithms are distinguished by the method to determine priority for selecting nodes.

\begin{algorithm}[H]
\caption{Basic Search Algorithm}
\label{algorithm_0}
\begin{algorithmic}[1]
\Function{Search}{$\mathcal{N}_0$}
    \State Initialize open list $\mathcal{L} \gets \{\mathcal{N}_0\}$
    \While{$\mathcal{L}$ is not empty}
        \State $\mathcal{N} \gets$ Pop a node from $\mathcal{L}$ based on priority
        \If{$\mathcal{N}$ is the destination}
            \State \Return $\mathcal{N}$
        \EndIf
        \State Expand $\mathcal{N}$ to obtain neighboring nodes
        \State Add neighboring nodes to $\mathcal{L}$
    \EndWhile
    \State \Return None
\EndFunction
\end{algorithmic}
\end{algorithm}


\noindent
\textbf{Depth-First Search (DFS)} prioritizes nodes with greater depth and explores as far as possible
before backtracking.\\ 
\noindent
\textbf{Breadth-First Search (BFS)} explores all neighbors at the current level before moving on to the next level.\\
\noindent
\textbf{Greedy Best First Search (GBFS)} uses a heuristic evaluation function $h(n)$ as the cost function, i.e., $f(n) = h(n)$. Here, $h(n)$ represents the cost from the current node to the destination. It can guide the search algorithm towards the destination but does not guarantee an optimal path.\\
\noindent
\textbf{Dijkstra's Algorithm} uses a movement cost function $g(n)$ as the cost function, i.e., $f(n) = g(n)$. Here, $g(n)$ represents the cost of moving from the starting point to the current node. It finds the shortest path from a starting node to all other nodes by considering the weights of edges.\\
\noindent
\textbf{A* Algorithm} combines the benefits of Dijkstra's Algorithm and GBFS. The cost function is defined as: $f(n) = g(n) + h(n)$.
It balances efficiency and optimality, making it highly effective for path planning.

\subsection{\textsc{AKeyS} Algorithm}

\begin{figure*}[t]
  \centering
   \includegraphics[width=0.9\linewidth]{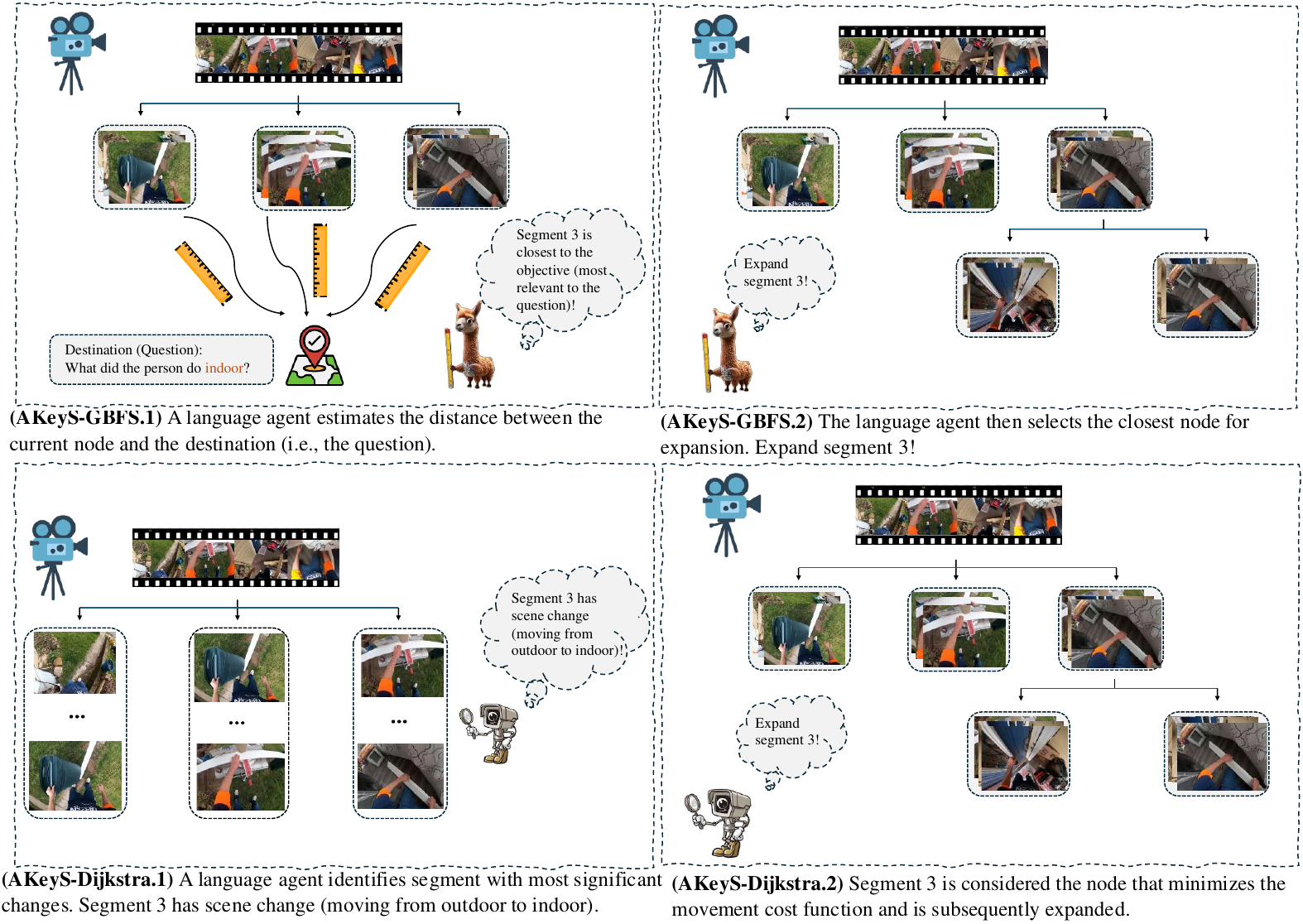}
    \caption{Illustration of \textsc{AKeyS}'s cost function evaluation and node expansion steps.}
   \label{fig:grid}
\end{figure*}

In the Method section, we first define the search objective, nodes, cost function, and termination conditions in our \textsc{AKeyS} algorithm, providing a comprehensive overview of the tree-structured keyframe search process. We also explain how the algorithm utilizes the retrieved information to answer questions. The key steps of \textsc{AKeyS} (leveraging language agents to evaluate the cost function and node expansion) are illustrated in Figure \ref{fig:grid}.

\noindent
\textbf{Search Objective}~~In \textsc{AKeyS}, keyframes are defined as frames containing key information about the question. The search objective is to identify a sufficient set of keyframes whose combined information is sufficient to answer the question. 
For example, humans can view only these keyframes instead of watching the entire video, to answer the question. 
When using MLLMs for VideoQA, we can also discard non-keyframes and adopt one of the following two approaches: (1) directly input the keyframes into an image-based MLLM to generate an answer, or (2) apply a VLM such as BLIP~\cite{li2022blipbootstrappinglanguageimagepretraining} to caption the keyframes, and use the captions to 
derive an answer. The two approaches are essentially the same, as they rely on the information within the keyframes and the learned priors of models.\\
\noindent
\textbf{Nodes}~~In \textsc{AKeyS} algorithm, we divide the video into multiple video segments, with each video segment representing a node. The initial node $\mathcal{N}_0$, is the entire video, which is first uniformly split into $M$ segments, where $M$ is a tunable hyper-parameter. These video segments are then put into an open list $\mathcal{L}$. The next node to be expanded, or the next video segment to be processed, is selected based on the cost function $f(n)$ we define. The expansion process means further subdividing the selected video segment. In this work, we perform a binary split on the segment for node expansion.\\
\noindent
\textbf{Answer Prediction}~~We define the first and last frames of all current video segments as \textbf{Visible Frames} $\mathcal{F}_{v}$. They are connected to each other, meaning that the last frame of one video segment is the first frame of the next. We can fully utilize the information in the visible frames, while the information in the other frames
is temporarily inaccessible. For the visible frames, we can employ either of the two approaches mentioned above: directly inputting the frames into the MLLM or first generating captions and then performing reasoning in the textual modality. In either way, we predict an answer based on the information from the visible frames. In this work, we choose the second approach. 
The predicted answer is a provisional guess during the intermediate stages and may change as the search progresses and more visible frames are revealed. When the termination condition is met, the search process concludes, and the predicted answer becomes the final answer. The total number of visible frames serves as a measure of the frame efficiency of the QA system: fewer visible frames mean fewer images for the MLLM to process, which results in higher efficiency. The final visible frames represent the keyframes obtained through our search process.\smallskip\\ 
\noindent
\textbf{Cost Function}~~Leveraging the evaluation capability of the language agents, we design different cost function based on different basic search algorithms.
\begin{itemize}
    \item \textbf{\textsc{AKeyS-GBFS}}~~In Greedy Best-First Search (GBFS), the cost function $h(n)$ represents the distance from the current node to the destination. Accordingly, we let the language agent evaluate the current visible frame's information and identify what visual information is missing for answering the question. The missing information can be seen as the distance between the current node and the destination. GBFS algorithm selects the node with the smallest $h(n)$ to expand. In our adaptation, the language agent attempts to identify \textbf{the missing visual information is likely located between which two specific invisible frames}, determining which video segments should be expanded. This leads to a variant of the \textsc{AKeyS} algorithm named \textsc{AKeyS-GBFS}.
    
    \item \textbf{\textsc{AKeyS-Dijkstra}}~~For Dijkstra's Algorithm, the cost function $g(n)$ represents the cost of moving from the start point to the current node. In our adaption, we have the language agent assess the current visible frame's information to identify \textbf{which video segment exhibits the most significant scene change} (e.g., the primary scenes, figures, or activities in the first and last frame of a video segment differ, indicating a transition or an important visual element's introduction). 
    Note that in Dijkstra's algorithm, the cost function does not consider the destination location, and similarly, in \textsc{AKeyS-Dijkstra}, the question is invisible to the language agent.\\
    We would like to further discuss why the video segment with the most significant scene change is considered the node closest to the start point. When segmenting and extracting keyframes from a long-form video with multiple scene transitions, the ideal scenario would be to treat each scene as a separate segment. This ensures that
    the video's visual elements are non-overlapping and non-missing in the visible frames, leading to the fewest visible frames needed and highest frame efficiency. Meanwhile, adjacent visible frames would contain information from two distinct scenes, allowing for visual comparison. This comparison would help QA system infer major changes in the video, leading to a better overall understanding of the video content. Therefore, intuitively, regardless of the specific question, treating each scene as an individual video segment is an optimal segmentation strategy. It is the most efficient segmentation method that achieves the same level of accuracy, which can be view as an abstraction of the shortest distance from the start point that achieves the same result.  Hence, we consider the video segment with the most significant scene change to be the node closest to the start point.
    \item \textbf{\textsc{AKeyS-A*}}~~For the A* Algorithm, the cost function is the sum of the heuristic evaluation function and the movement cost function, i.e., $f(n) = h(n) + g(n)$, which means that A* Algorithm takes into account both the distance from the current node to the destination and the distance from the start point to the current node. Correspondingly, in our \textsc{AKeyS-A*} variant, the language agent must simultaneously consider two factors: (1) which video segment is likely to contain the missing information, and (2) which video segment exhibits the most significant scene change. Only video segments that satisfy both are prioritized for expansion.
    \item \textbf{\textsc{AKeyS-BFS}}~~We also propose a naive algorithm variant, \textsc{AKeyS-BFS} which does not rely on a language agent to evaluate the cost function. Instead, it performs a breadth-first expansion, continually splitting all the existing video segments (in the case of no pruning). Like BFS, \textsc{AKeyS-BFS} advances in a wave-like manner, steadily progressing. This variant is suitable for situations where a language agent cannot be accessed, or where the overhead introduced by the LLM is less of a concern, with a greater emphasis on ensuring no information is overlooked.
\end{itemize}
We do not intend to introduce an \textsc{AKeyS-DFS} variant, as its depth-first expansion focuses on a single initial segment. Without strict termination conditions, it is prone to falling into local optima.\smallskip\\
\textbf{Termination Condition}~~Traditional search algorithms typically have a deterministic termination condition: whether the search objective has been reached. However, in keyframe search algorithms for VideoQA, the termination condition is much more vague and difficult to define. It is challenging to determine whether sufficient information has been gathered, or key information is missing or over-inference. Inspired by the reflection, summarization, and self-evaluation abilities of language agents, we use the base LLM to evaluate the confidence in the predicted answer and determines whether to terminate the search accordingly. In this way, the \textsc{AKeyS} will terminate when a sufficiently confident prediction is made. Specifically, we combine two methods of confidence evaluation by a voting mechanism, as outlined below.

\begin{algorithm}[t]
\caption{Agentic KeyFrame Search (\textsc{AKeyS})}
\label{algorithm_1}
\begin{algorithmic}[1]
\Require Video $v$, question $q$, MLLM $F$, confidence threshold $C$, max iteration $T$, uniform sampling size $M$, beam size $B$
\Ensure Answer $\hat{y}$, keyframes $\{\mathcal{F}_k\}$
\State $\mathcal{N}_0 \gets v$
\State $\mathcal{N}_1, \mathcal{N}_2, ... ,\mathcal{N}_{M}\gets \text{UniformSegment}(\mathcal{N}_0, M)$
\State $\mathcal{L} \gets \{\mathcal{N}_0, \mathcal{N}_1, ... , \mathcal{N}_M\}$
\State $t \gets 1$
\While{$t \leq T$}
    \State $\{F_v\} \gets \text{ExtractVisibleFrames}(\mathcal{L})$
    \State $\hat{y} \gets \text{PredictAnswer}(F, \{\mathcal{F}_v\}, q)$
    \State $c_1, c_2 \gets \text{EvaluateConfidence}(F, \hat{y}, \{\mathcal{F}_v\}, q)$
    \If{$c_1 \geq C\, \text{and}\, c_2 \geq C$}
        \State \textbf{break} 
    \Else
        \State $\{p\} \gets \text{EvaluateCostFunction}(F, \{\mathcal{F}_v\}, \mathcal{L})$
        \State $\mathcal{L} \gets \text{SelectAndExpandNodes}(\mathcal{L}, \{p\}, B)$
    \EndIf
    \State $t \gets t+1$
\EndWhile
\State $\{\mathcal{F}_k\} \gets \{\mathcal{F}_v\}$
\State \Return $\hat{y}$, $\{\mathcal{F}_k\}$
\end{algorithmic}
\end{algorithm}

\begin{table*}[ht!]
\caption{\textbf{Comparison between \textsc{AKeyS} and other methods.} We highlight the gain of our method over VideoTree~\cite{wang2024videotreeadaptivetreebasedvideo} in blue.}
\centering
\setlength{\tabcolsep}{1.4mm}
\begin{tabular}{lccccccc}
\toprule
\textbf{Model} & \textbf{(M)LLM} & \multicolumn{2}{c}{\textbf{EgoSchema}} & \multicolumn{4}{c}{\textbf{NExT-QA}} \\
\cmidrule(lr){3-4} \cmidrule(lr){5-8}
 & & \textbf{Sub.} & \textbf{Full} & \textbf{Tem.} & \textbf{Cau.} & \textbf{Des.} & \textbf{Avg.} \\
\midrule
\multicolumn{8}{c}{\textit{Based on Open-source Captioners and LLMs}} \\
MVU~\cite{ranasinghe2025understandinglongvideosmultimodal}& Mistral-13B & 60.3 & 37.6 & 55.4 & 48.1 & 64.1 & 55.2 \\
LangRepo~\cite{kahatapitiya2024languagerepositorylongvideo} & Mixtral-8x7B & 66.2 & 41.2 & 51.4 & 64.4 & 69.1 & 60.9 \\
Video-LLA+INTP~\cite{shang2024interpolatingvideollmslongersequencelmms} & Vicuna-7B v1.5 & - & 38.6 & 58.6 & 61.9 & 72.2 & 62.7 \\
\midrule
\multicolumn{8}{c}{\textit{Based on Proprietary MLLMs}} \\

IG-VLM~\cite{kim2024imagegridworthvideo} & GPT-4V & 59.8 & - & 63.6 & 69.8 & 74.7 & 68.6 \\
LVNet~\cite{park2024framesusefulefficientstrategies} & GPT-4o & 68.2 & 61.1 & 65.5 & 75.0 & 81.5 & 72.9 \\
\midrule
\multicolumn{8}{c}{\textit{Based on Open-source Captioners and Proprietary LLMs}} \\
ProViQ~\cite{choudhury2023zeroshotvideoquestionanswering} & GPT-3.5 & 57.1 & - & - & - & - & 64.6 \\
MoReVQA~\cite{min2024morevqaexploringmodularreasoning}& PaLM-2 & - & 51.7 & 64.6 & 70.2 & - & 69.2 \\
Vamos~\cite{wang2024vamosversatileactionmodels} & GPT-4 & 51.2 & 48.3 & - & - & - & - \\
LLoVi~\cite{zhang2024simplellmframeworklongrange} & GPT-4 & 61.2 & - & 61.0 & 69.5 & 75.6 & 67.7 \\
VideoAgent~\cite{wang2024videoagentlongformvideounderstanding} & GPT-4 & 60.2 & 54.1 & 64.5 & 72.7 & 81.1 & 71.3 \\
VideoAgent~\cite{fan2024videoagentmemoryaugmentedmultimodalagent} & GPT-4 & 62.8 & 60.2 & - & - & - & - \\
LifelongMemory~\cite{wang2024lifelongmemoryleveragingllmsanswering} & GPT-4 & 64.1 & 58.6 & - & - & - & - \\
VideoTree~\cite{wang2024videotreeadaptivetreebasedvideo} & GPT-4 & 66.2 & 61.1 & 70.6 & 76.5 & 83.9 & 75.6 \\
\midrule
\textsc{AKeyS} (Ours) & GPT-4 & \textbf{68.0 (\textcolor{blue}{1.8 $\uparrow$})} & \textbf{63.1 (\textcolor{blue}{2.0 $\uparrow$})} & \textbf{72.3 (\textcolor{blue}{1.7 $\uparrow$})} & \textbf{78.2 (\textcolor{blue}{1.7 $\uparrow$})} & \textbf{85.4 (\textcolor{blue}{1.5 $\uparrow$})} & \textbf{77.4 (\textcolor{blue}{1.8 $\uparrow$)}} \\
\textsc{AKeyS} (Ours) & GPT-4o & \textbf{68.6 (\textcolor{blue}{2.4 $\uparrow$})} & \textbf{63.6 (\textcolor{blue}{2.5 $\uparrow$})} & \textbf{72.9 (\textcolor{blue}{2.3 $\uparrow$})} & \textbf{79.0 (\textcolor{blue}{2.5 $\uparrow$})} & \textbf{86.1 (\textcolor{blue}{2.2 $\uparrow$})} & \textbf{78.1 (\textcolor{blue}{2.5 $\uparrow$)}} \\
\bottomrule
\end{tabular}
\label{table:main_results}
\end{table*}

\begin{itemize}
    \item \textbf{Self-Evaluation and Self-Reflection}~~
    LLMs can be instructed to self-evaluate their responses, reflecting on potential shortcomings in their responses~\cite{shinn2023reflexionlanguageagentsverbal, ren2023selfevaluationimprovesselectivegeneration}. Therefore, after generating an answer, we input the question, information of visible frames, and the LLM's previous reasoning chain and predicted answer back into the model. 
    The LLM then assesses the accuracy and reliability of its previous answer and output a confidence score ($c_1$).
    
    \item 
    \textbf{Temporal Summarization}~~
    The captions of the sampled frames are discrete.
    To integrate the sampled frames along the temporal dimension, we instruct the LLM to summarize their captions to form a cohesive overview of the video.
    We use few-shot examples~\cite{brown2020languagemodelsfewshotlearners} to generate a more accurate and detailed video summary. Then we prompt the LLM to predict the answer and output a confidence score ($c_2$) based on the summary. The advantage of this approach is to consider the sampled frames in a complete temporal context rather than in isolation.
\end{itemize}
We employ a voting mechanism to ensemble the above two methods. The search process only terminates when both methods independently determine that they have sufficient confidence ($c_1 \geq C\, \text{and}\, c_2 \geq C$, $C$ is the threshold).

Other search techniques can also be integrated into \textsc{AKeyS} framework, for instance, using beam search to expand multiple nodes each step. 
It not only reduces the computational overhead of evaluating the cost function by language agents, but also helps prevent the search algorithm from getting stuck in local optima, as shown in Algorithm \ref{algorithm_1}. 

Finally, we would like to comment that the term \textbf{agentic} in our algorithm's name reflects in two aspects: (1) the use of an LLM to evaluate the cost function, which enables the LLM to engage in path planning while interacting with the environment (i.e., the video and the question); (2) the use of the LLM to assess the termination condition, which enables the LLM to engage in decision-making. These two aspects together endow the base LLM with agentic properties, making it clear that \textsc{AKeyS} is essentially powered by a language agent.

\section{Experiments}
\label{sec:experimentsa}

\subsection{Datasets}
\noindent
{\bf EgoSchema}~\cite{mangalam2023egoschemadiagnosticbenchmarklongform}  dataset comprises over 5,000 human-curated multiple-choice question-answer pairs, making it one of the most widely used datasets for long-form video question and answering. Its subset contains 500 video and QA pairs. 
Each video in the datsset is three minutes in length. 
A notable feature of EgoSchema is its high difficulty level: humans can only achieve 76\% accuracy, and current Video-LLMs perform below 70\%. The extended video length and increased complexity underscore the importance of keyframe search and key information retrieval.

\noindent
{\bf NExT-QA}~\cite{xiao2021nextqanextphasequestionansweringexplaining} dataset consists of 5,440 videos and approximately 52K manually annotated question-answer pairs. Its primary focus is to assess whether QA models truly understand the causal and temporal structures of actions within a video.
We use the multiple-choice QA part of NExT-QA. Based on the types, the questions are divided into casual questions, temporal questions and descriptive questions.






\subsection{Main Results}


Following VideoTree~\cite{wang2024videotreeadaptivetreebasedvideo}, we compare the performance of \textsc{AKeyS} with various related approaches on LLM-driven VideoQA using \textsc{AKeyS-A*} variant. Most of the baselines are mentioned in the related work and Appendix Section A. Implementation details are provided in Appendix Section B. Prompts we use are listed in Appendix Section C. Table \ref{table:main_results} demonstrates that \textsc{AKeyS} significantly outperforms all these baselines. Specifically, \textsc{AKeyS} (with GPT-4 as base LLM) achieves 63.1\% accuracy on EgoSchema fullset (surpassing the best baseline by 2.0\%) and 77.4\% accuracy on NExT-QA (surpassing the best baseline by 1.8\%).
Moreover, \textsc{AKeyS} operates in a training-free, zero-shot setting, while it still outperforms training-based methods such as LVNet~\cite{park2024framesusefulefficientstrategies} and Vamos~\cite{wang2024vamosversatileactionmodels}. Meanwhile, \textsc{AKeyS} processes only visible frames, for instance, achieving the reported performance requires only about 15\% of the total frames. In contrast, methods like LangRepo~\cite{kahatapitiya2024languagerepositorylongvideo} and LifeLongMemory~\cite{wang2024lifelongmemoryleveragingllmsanswering} process all frames without selection. 

Additionally, as shown in Figure \ref{fig:frame_efficiency}, we compare \textsc{AKeyS}'s frame efficiency with other keyframe extraction methods in the same condition. The results show that \textsc{AKeyS} utilizes frames more efficiently than LLoVi~\cite{zhang2024simplellmframeworklongrange}, VideoAgent~\cite{wang2024videoagentlongformvideounderstanding} and VideoTree~\cite{wang2024videotreeadaptivetreebasedvideo}, demonstrating its superior ability to identify key information.



\subsection{Ablation Studies}

\subsubsection{Basic Search Algorithms}

In Table \ref{tab:search_algorithm}, We investigate performance and frame efficiency of several \textsc{AKeyS} algorithm variants with different base search algorithms on the EgoSchema subset. The frame efficiency is measured by the number of visible frames\footnote{On EgoSchema dataset, all videos are three minutes long, and we set the frame rate $fps=1$, which means the overall frame number is 180.}. We observe that \textsc{AKeyS-A*} achieves the highest accuracy. \textsc{AKeyS-BFS} ranks second in accuracy but has lower frame efficiency, as BFS Algorithm exhaustively explores all branches, leading to higher exploration costs. \textsc{AKeyS-GBFS} slightly outperforms \textsc{AKeyS-Dijkstra} on both metrics, while \textsc{AKeyS-A*} combines the strengths of them, significantly improving accuracy with only a slight compromise in frame efficiency. This demonstrates that efficient keyframe localization requires both the heuristic search function and the movement cost function.

\subsubsection{Termination Condition}

In Table \ref{tab:termination_condition}, we conduct ablation experiments on the termination condition of the search process, using \textsc{AKeyS-A*} on the EgoSchema subset. The results show that self-evaluation \& self-reflection and temporal summarization assess information sufficiency from different perspectives. When combined, they enhance the reliability of confidence estimation, leading to improved algorithm performance.

\subsubsection{Base LLM}

We also conducted an ablation study on the base LLM of the \textsc{AKeyS} algorithm in Table \ref{tab:base_llm}. We find that GPT-4o achieves the best performance as the base LLM. In contrast, reasoning models such as o3-mini and Deepseek-R1 perform slightly worse than GPT-4o, likely due to the relatively straightforward nature of visual reasoning in our tasks.

\begin{table}
\caption{\textbf{Ablation on basic search algorithms}. We highlight the improvement of \textsc{AKeyS-A*} over the naive \textsc{AKeyS-BFS} in the table, emphasizing the role of the cost function evaluation.}
  \centering
  \setlength{\tabcolsep}{3.0mm}
  \begin{tabular}{@{}lcc@{}}
    \toprule
    \textbf{Algorithm} & \textbf{Accuracy} &  \textbf{\# Visible Frames}\\
    \midrule
    \textsc{AKeyS-BFS} & 64.7 & 31.2 \\
    \textsc{AKeyS-GBFS} & 67.0 & \textbf{27.3} \\
    \textsc{AKeyS-Dijkstra} & 66.8 & 27.6 \\
    \textsc{AKeyS-A*} & \textbf{68.0 (\textcolor{blue}{3.3 $\uparrow$})} & 27.9 \\
    \bottomrule
  \end{tabular}
  \label{tab:search_algorithm}
\end{table}


\begin{table}
\caption{\textbf{Ablation on termination condition}}
  \centering
      \setlength{\tabcolsep}{0.9mm}
  \begin{tabular}{@{}lcc@{}}
    \toprule
    \textbf{Termination Condition} & \textbf{Accuracy} & \textbf{\# Visible Frames}\\
    \midrule
    Self-Evaluation & 67.4 & \textbf{27.4} \\
    Summarization & 67.3 & 28.2 \\
    Vote & \textbf{68.0} & 27.9 \\
    \bottomrule
  \end{tabular}

  \label{tab:termination_condition}
\end{table}


\begin{table}
  \caption{\textbf{Ablation on different base LLMs}}  \label{tab:base_llm}
  \centering
    \setlength{\tabcolsep}{3.9mm}
  \begin{tabular}{@{}lcc@{}}
    \toprule
    \textbf{Base LLM} & \textbf{Accuracy} & \textbf{\# Visible Frames}\\
    \midrule
    \textsc{GPT-4} & 68.0 & 27.9 \\
    \textsc{GPT-4o} & \textbf{68.6} & \textbf{26.7} \\
    \textsc{o3-mini} & 67.3 & 28.3 \\
    \textsc{Deepseek-R1} & 67.6 & 26.9 \\
    \textsc{LLaMA-3.3-70B} & 65.2 & 27.4 \\
    \bottomrule
  \end{tabular}

\end{table}


\begin{table*}[ht!]
\begin{center}
\begin{threeparttable}[b]
\caption{\textbf{Comparison of computation costs between Video-LLMs and \textsc{AKeyS}}}
\setlength{\tabcolsep}{1.6mm}
\begin{tabular}{lccccc}
\toprule

\textbf{Model} & \textbf{Train Model Size} & \textbf{Tain Data Size} & \textbf{Computational Resources} & \textbf{EgoSchema} & \textbf{NExT-QA} \\

\midrule

ViLA~\cite{wang2024vilaefficientvideolanguagealignment} & 4B & 36.4K & 8 $\times$ 40 GB A100s & - & 75.6\ \\

VideoChat2~\cite{li2024mvbenchcomprehensivemultimodalvideo}& 7B & 4M & 32 $\times$ 80 GB A100s & 54.4 & 78.6  \\

VideoLLaMA2~\cite{cheng2024videollama2advancingspatialtemporal} & 72B & 13.6M & 32 $\times$ 80 GB A100s & 63.9 & 75.6 \\

\midrule

\textsc{AKeyS} (ours) &  & Training-free &  & 63.1 & 77.4  \\

\bottomrule
\end{tabular}

\label{table:compare_with_vlm}

\end{threeparttable}
\end{center}
\end{table*}

\section{Analysis}

\begin{figure*}[t]
  \centering
   \includegraphics[width=0.9\linewidth]{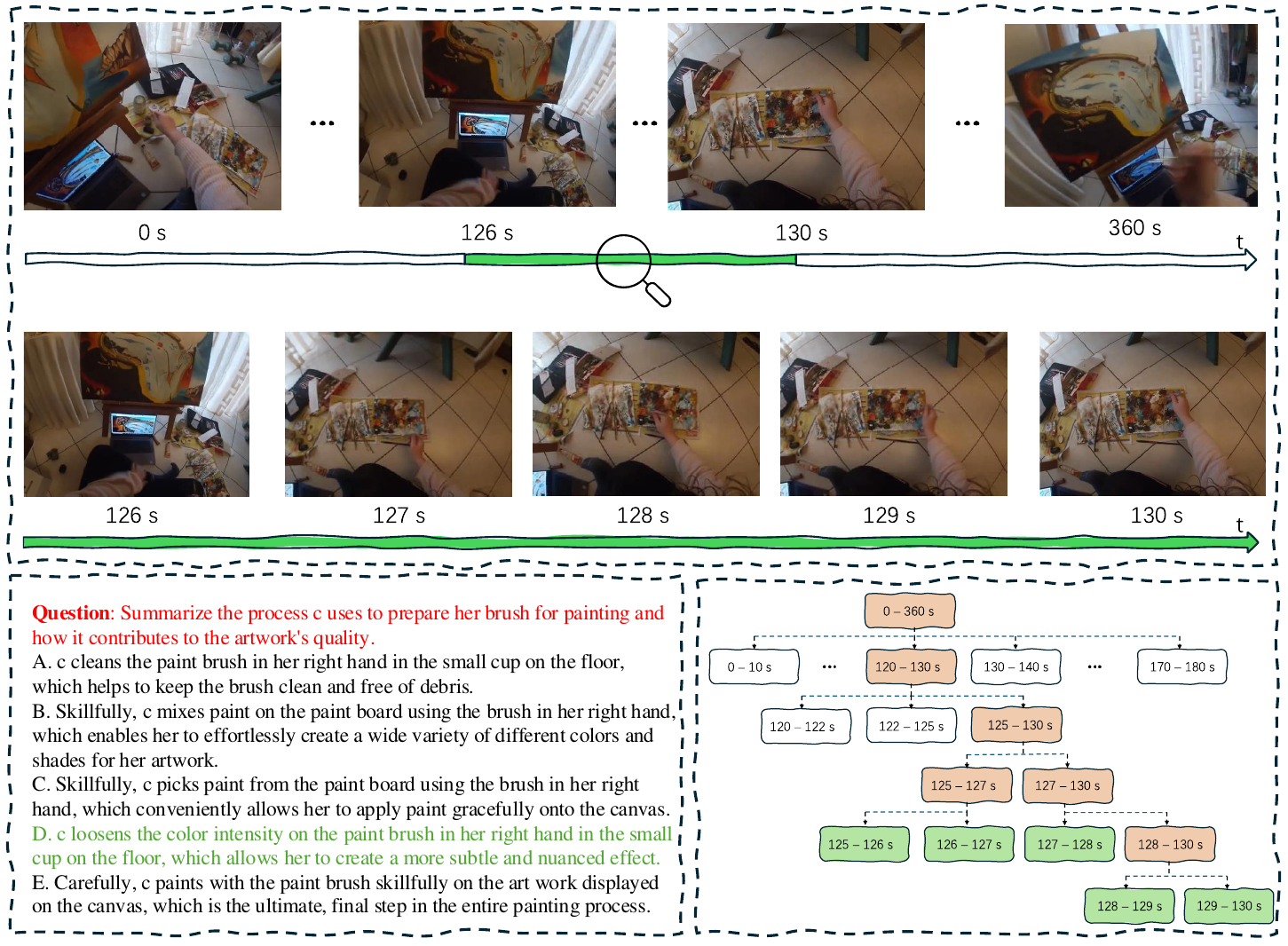}
   \caption{Visualization of tree-search process of a case from EgoSchema~\cite{mangalam2023egoschemadiagnosticbenchmarklongform}.}
   \label{fig:visualization}
\end{figure*}


\subsection{Comparison with Video-LLMs}

As previously discussed, there are two primary method for VideoQA task: (I) utilizing Video-LLMs for end-to-end computation; (II) employing (M)LLM-driven, keyframe-based, training-free method, like \textsc{AKeyS}. We argue that both methods have their respective advantages. The key strength of Method I is that state-of-the-art Video-LLMs~\cite{gao2024linvtempowerimagelevellarge, chen2025expandingperformanceboundariesopensource} outperform Method II. It is suitable for scenarios where high accuracy is required, and computational cost is not a concern. In contrast, the primary advantage of Method II is its practical value for daily video analysis tasks, as it offers a more favorable balance between performance and computational cost. In the following, we use \textsc{AKeyS} as an example to illustrate the relative advantages of Method II.

\begin{itemize}
    \item \textbf{Training-Free}. The training-free nature of Method II significantly reduces the overall cost. In Table \ref{table:compare_with_vlm}, we present the training costs and resource requirements of Video-LLMs that achieve comparable performance with \textsc{AKeyS} on EgoSchema~\cite{mangalam2023egoschemadiagnosticbenchmarklongform} and NExT-QA~\cite{xiao2021nextqanextphasequestionansweringexplaining} benchmarks. The table highlights the complexity and high cost of training Video-LLMs, underscoring the training-free advantage of Method II.

    \item \textbf{Lower Inference Overhead}. 
    Method II still relies on large model inference. However, \textsc{AKeyS} significantly reduces inference overhead by efficient keyframe selection instead of processing the whole video.

    \item \textbf{Better Interpretability}. 
    \textsc{AKeyS} provides greater interpretability by generating intermediate results, such as the keyframe selection and textual reasoning process. In contrast to the end-to-end nature of Method I, this enhances transparency and interpretability. 
\end{itemize}


\subsection{Visualization}


Figure \ref{fig:visualization} presents a visualized case study. In this 3-minute video, the key information for answering the question is located between 126s and 130s. Our \textsc{AKeyS} algorithm precisely identifies this critical video segment by searching along the video tree and expanding relevant nodes. And it retrieves all frames within the 125s–130s range as keyframes, successfully answering the question. 
In the video tree, we mark the nodes traversed by the key search path in yellow and mark the final leaf node obtained from the key search path in green. The nodes outside the key search path are barely expanded.

\section{Conclusion}



In this paper, we introduce \textsc{AKeyS}, a novel keyframe search algorithm tailored for efficient video analysis. \textsc{AKeyS} leverages the language agent to guide the search process. 
Like separating the wheat from the chaff, it effectively distinguishes key information from redundancy in videos.
We evaluate \textsc{AKeyS} on the EgoSchema and NExT-QA datasets, where it achieves higher accuracy and frame efficiency than all baseline methods. 
We believe that \textsc{AKeyS} represents a significant step towards building more powerful video agents and tackling various video understanding challenges.
{
    \small
    \bibliographystyle{ieeenat_fullname}
    \bibliography{ref}
}



\end{document}